\documentclass[]{seed_like}

\usepackage{graphicx}
\usepackage{booktabs}      
\usepackage{multirow}      
\usepackage{multicol}      
\usepackage{arydshln}      
\usepackage{adjustbox}     
\usepackage{makecell}      
\usepackage{colortbl}      
\usepackage{orcidlink}
\usepackage{amsmath, amsfonts, amssymb}
\usepackage{url}
\usepackage{booktabs, multirow, arydshln, adjustbox, makecell}
\usepackage{pifont, microtype, nicefrac, xcolor}
\usepackage{enumitem, wrapfig, algorithm, algpseudocode}
\usepackage{xspace}

\definecolor{lightblue}{rgb}{0.796, 0.894, 0.9808}
\definecolor{naturegreen}{RGB}{0,102,85}
\definecolor{cvprblue}{rgb}{0.21,0.49,0.74}
\definecolor{lightblue}{rgb}{0.796, 0.894, 0.9808}
\definecolor{forestgreen}{RGB}{34,139,34}
\definecolor{crimson}{RGB}{220,20,60}
\definecolor{wine}{HTML}{830E0D}

\newtheorem{definition}{Definition}[section]

\title{Staying \texttt{VIGIL}ant: Mitigating Visual Laziness via Counterfactual Visual Alignment in MLLMs}

\author[1,*]{Xi Xiao}
\author[2,*]{Chen Liu}
\author[3]{Chih-Ting Liao}
\author[4]{Yunbei Zhang}
\author[1]{Qizhen Lan}
\author[5]{Yuxiang Wei}
\author[6]{Lin Zhao}
\author[4]{Janet Wang}
\author[7,\dag]{Jianyang Gu}
\author[8]{Muchao Ye}
\author[1,\dag]{Tianyang Wang}
\author[9]{Hao Xu}

\affiliation[1]{UAB}
\affiliation[2]{Yale}
\affiliation[3]{UNSW}
\affiliation[4]{Tulane}
\affiliation[5]{Georgia Tech}
\affiliation[6]{NEU}
\affiliation[7]{OSU}
\affiliation[8]{UIowa}
\affiliation[9]{Harvard}

\contribution[*]{Equal Contributions}
\contribution[\dag]{Co-advising}

\abstract{
Multimodal large language models~(MLLMs) extend large language models~(LLMs) with visual perception, enabling joint reasoning over images and text. Despite inheriting strong reasoning capabilities from LLMs, they remain prone to hallucinations that contradict their visual inputs. Mechanistic studies indicate that this weakness stems from \textit{visual laziness}: MLLMs encode the correct visual evidence internally, but overly rely on strong language priors during response.
Existing alignment methods, such as direct preference optimization, primarily optimize outcome-level rewards based on text. This introduces an optimization bias toward linguistic shortcuts, leading to responses that often contradict the visual evidence. To address this, we propose \underline{V}isual \underline{I}nformation \underline{G}ain \underline{I}n a\underline{L}ignment (\texttt{VIGIL}), a reinforcement-learning~(RL) post-training framework that shifts the focus from numerical reward fitting to causal visual grounding. \texttt{VIGIL} introduces a geometric constraint that explicitly maximizes the mutual information between the visual input and the generated response. We achieve this by penalizing ``blind confidence'' instances where the model remains improperly certain even when textual-visual attention is masked to create a counterfactual blind state. Extensive experiments show that \texttt{VIGIL} consistently outperforms recent alignment methods across hallucination and reasoning benchmarks without compromising text-only capabilities. Our approach matches the full-data performance of state-of-the-art methods using only 25\% of the preference data and even demonstrates emergent spatial grounding capabilities without explicit bounding box supervision.

}

\correspondence{Xi Xiao (\email{xixiao@uab.edu})}

\checkdata[Project Page]{\url{https://xixiaouab.github.io/VIGIL/}}

\begin{document}
\maketitle

\makeatletter
\let\old@vspace\@vspace
\let\old@vspacer\@vspacer
\renewcommand{\@vspace}[1]{}
\renewcommand{\@vspacer}[1]{}
\makeatother

\section{Introduction}
\label{sec:intro}
\vspace{-4pt}

Multimodal large language models (MLLMs) are transitioning from basic perception tools to sophisticated reasoning engines capable of interpreting complex visual data~\cite{zhang2025system}. Modern architectures, such as Qwen2-VL~\cite{bai2023qwen}, LLaVA-OneVision~\cite{li2024llava} and InternVL2.5~\cite{chen2024internvl}, can conduct complex reasoning, including solving mathematical problems and analyzing financial charts. However, \textit{hallucination} remains a persistent barrier to their reliability. In the multimodal context, hallucination is not simply a perceptual error, but rather represents a failure of visual grounding where models generate descriptions that lack support from the physical visual input.

The root cause of this unreliability is often \textit{visual laziness}~\cite{zhang2025mllms}. Modern MLLMs typically combine a visual encoder with a pre-trained, text-centric LLM. Because the LLM component is trained on vast text corpora, it harbors strong \textit{language priors}. When evaluating complex multimodal inputs, the model’s reasoning often defaults to these inherent priors rather than relying on visual evidence. Recent studies~\cite{zhang2025mllms, leng2024mitigating, long2026understanding, bai2025hallucination, zhu2026leveraging} reveal a striking disparity: MLLMs often contain the correct visual features in their latent space, but still generate incorrect answers because linguistic correlations hijack the decoding process.

Existing approaches to mitigate this issue have clear trade-offs. One direction relies on architectural modifications, such as the introduction of external tools or modular designs~\cite{kang2026vgent} to separate reasoning from perception, which complicates end-to-end learning. Another direction uses online reinforcement learning for dynamic visual token selection~\cite{lin2026adaptvision}, introducing significant computational overhead and training instability. Meanwhile, standard outcome-based alignment methods, such as direct preference optimization~(DPO)~\cite{rafailov2023direct} and its difficulty-aware variants~\cite{qiu2025dadpo}, optimize the policy by minimizing a preference loss on the final text output. They penalize incorrect tokens but do not constrain how the model derives its answer. If MLLMs arrive at the correct answer through language priors rather than visual evidence, standard DPO still rewards the output, inadvertently reinforcing these optimization shortcuts and preserving visual laziness.
\\



Taken together, we argue that building robust multimodal alignment and reducing visual laziness requires explicitly enforcing reliance on \textit{visual evidence} rather than \textit{language priors}. Our core intuition is simple: to teach a model to use visual information, we must expose the consequences of being blind. Motivated by this insight, we introduce \underline{V}isual \underline{I}nformation \underline{G}ain \underline{I}n a\underline{L}ignment (\texttt{VIGIL}), a post-training framework that maximizes the Visual Information Gain (VIG) between the visual input and the generated response.

Specifically, during optimization, we construct a counterfactual blind state~($x^\emptyset_\text{v}$) by masking visual attention, preventing the model from accessing visual evidence. The policy is then optimized to maximize the divergence between the standard seeing state and the introduced counterfactual blind state. This divergence acts as a geometric constraint on the model’s response distribution, penalizing blind confidence cases where the model remains confident even without visual input. As a result, \texttt{VIGIL} separates visually grounded reasoning from predictions driven purely by language priors, ensuring that high-confidence outputs are causally anchored to visual features~($x_\text{v}$).

Our contributions are summarized as follows:

\begin{enumerate}[leftmargin=16pt, topsep=0pt, itemsep=4pt, parsep=0pt]
    \item \textbf{A counterfactual alignment framework:} We propose \texttt{VIGIL} to mitigate multimodal hallucinations by incorporating counterfactual visual decoupling into the preference optimization loop and explicitly maximizing the Visual Information Gain.
    
    \item \textbf{Data and computational efficiency:} Operating entirely offline, \texttt{VIGIL} is stable and computationally lightweight. It matches the performance of competitive baselines using only 25\% of their preference data.
    
    \item \textbf{Consistent gains across scales and architectures:} Across diverse architectures ranging from 7B to 72B parameters, \texttt{VIGIL} consistently improves performance, with gains increasing for larger and stronger foundation models.
    
    \item \textbf{Emergent spatial grounding:} Without any explicit bounding box supervision, \texttt{VIGIL} improves zero-shot referring expression comprehension, indicating that spatial grounding can emerge without explicit localization supervision.
\end{enumerate}

{
\makeatletter
\let\oldresizebox\resizebox
\def\resizebox#1#2#3{%
  \strip@dollar#3\@nil
}
\def\strip@dollar$#1$\@nil{#1}

\section{Preliminaries}
\label{sec:preliminaries}
\vspace{-4pt}


In this section, we provide an extensive theoretical analysis of visual hallucination in MLLMs, which originates from a phenomenon we term \textit{visual laziness}.


\subsection{Definition of Visual Laziness}
\vspace{-4pt}
\label{subsec:mechanism}

Modern MLLMs often combine a visual encoder with a pre-trained LLM. During reasoning, although the visual encoder captures rich spatial features, an MLLM often overly relies on the inherent \textit{language priors} learned during massive text-only pre-training~\cite{zhang2025mllms}. This phenomenon, which we term \textit{visual laziness}, causes models to favor outputs using exclusively textual information despite having access to relevant visual clues.

\subsection{Information-Theoretic Analysis of Visual Laziness}
\vspace{-4pt}
\label{subsec:analysis}

Give a multimodal preference dataset $\mathcal{D} = \{(x_\text{v}, x_\text{t}, y_w, y_l)\}$, where $x_\text{v}$ is the visual input, $x_\text{t}$ is the textual instruction, and $(y_w, y_l)$ are the preferred and disfavored responses ($w$ and $l$ respectively stands for ``winning'' and ``losing''), standard Direct Preference Optimization (DPO)~\cite{rafailov2023direct} optimizes the policy $\pi_\theta$ by minimizing the negative log-likelihood of the preference:
\vspace{-4pt}
\begin{equation}
\label{eq:dpo_pre}
\resizebox{0.92\textwidth}{!}{$
\mathcal{L}_{\text{DPO}}(\pi_\theta; \pi_{\text{ref}}) = -\mathbb{E}_{\mathcal{D}} \left[ \log \sigma \left( \beta \log \frac{\pi_\theta(y_w | x_\text{v}, x_\text{t})}{\pi_{\text{ref}}(y_w | x_\text{v}, x_\text{t})} - \beta \log \frac{\pi_\theta(y_l | x_\text{v}, x_\text{t})}{\pi_{\text{ref}}(y_l | x_\text{v}, x_\text{t})} \right) \right]
$}
\vspace{-4pt}
\end{equation}
where $\pi_{\text{ref}}$ serves as the reference policy to prevent excessive drift.
We argue that the objective in Eq.~\ref{eq:dpo_pre} allows for an \textit{optimization shortcut} that exacerbates visual laziness. To understand this, we analyze the model's behavior through the lens of conditional probability and information redundancy.

\begin{figure*}[!t]
    \centering
    \includegraphics[width=\textwidth]{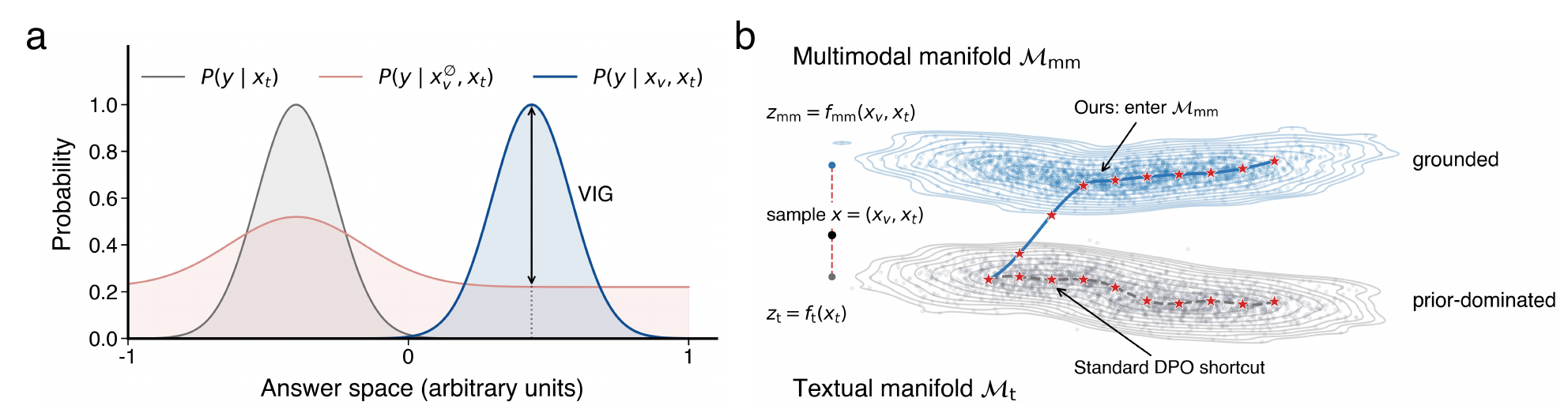}
    \caption{\textbf{Visual laziness in MLLMs.}
    \textbf{(a)}~Degenerate posterior and vanishing visual gain. When predictions are dominated by language priors $P(y | x_\text{t})$, the multimodal posterior $P(y | x_\text{v}, x_\text{t})$ collapses toward the vision-blind prior $P(y | x_\text{v}^{\emptyset}, x_\text{t})$, yielding a diminished visual information gain~(VIG).
    \textbf{(b)}~Optimization geometry of visual shortcuts. A geometric view illustrates an optimization shortcut that stays on the textual manifold $\mathcal{M}_\text{t}$ and converges to a prior-dominated region, instead of entering the multimodal manifold $\mathcal{M}_\text{mm}$ for grounded reasoning, motivating our \texttt{VIGIL} framework. Refer to~\cite{meilua2024manifold, liao2025rnagenscape, li2026back, liao2024assessing, sun2025geometry} for the definition of manifold.}
    \label{fig:prelim_visual_laziness}
\vspace{-10pt}
\end{figure*}

\paragraph{Language Prior Dominance.} 
\vspace{-4pt}
A hallucination occurs when the model's reasoning is dominated by the unimodal prior $P(y | x_\text{t})$ rather than the multimodal posterior $P(y | x_\text{v}, x_\text{t})$. If a preferred response $y_w$ can be inferred from the text alone, such as in instances of strong linguistic correlation, the model can minimize the loss by simply reinforcing its \textit{language priors} without truly grounding the answer in the visual content. Mathematically, this manifests as an approximation:
\vspace{-4pt}
\begin{equation}
\pi_\theta(y | x_\text{v}, x_\text{t}) \approx \pi_\theta(y | x_\text{t})
\vspace{-4pt}
\end{equation}
In this regime, the visual input $x_\text{v}$ becomes statistically redundant in the optimization trajectory. As illustrated in Fig.~\ref{fig:prelim_visual_laziness}, visual laziness manifests itself as both a \underline{degenerate posterior} with vanishing VIG (Fig.~\ref{fig:prelim_visual_laziness}a, see formal definition in Section~\ref{subsec:theory}) and an \underline{optimization shortcut} that stays in the textual manifold (Fig.~\ref{fig:prelim_visual_laziness}b).

\paragraph{Perception-Reasoning Decoupling.}
\vspace{-4pt}
When visual evidence $x_\text{v}$ is treated as redundant, the model's reasoning foundation collapses. For example, a model might predict that an apple is \underline{red} based on common textual associations even when the image depicts a \underline{green} apple, illustrating how a strong \textit{language prior} could override the visual signal, resulting in a decoupling between perception and reasoning. We argue that this failure mode reflects a fundamental limitation of existing multimodal alignment methods. While prior approaches have improved performance through sample reweighting~\cite{qiu2025dadpo} or increased architectural complexity~\cite{kang2026vgent}, they do not explicitly encourage the model to rely on visual evidence when forming its predictions. This observation motivates our proposed \texttt{VIGIL} framework. We posit that robust grounding requires maximizing the information contributed specifically by $x_\text{v}$, thereby decoupling the multimodal manifold $\mathcal{M}_\text{mm}$ from the pure textual manifold $\mathcal{M}_\text{t}$.

\subsection{Outcome Rewards vs. Dependency Alignment}

\vspace{-4pt}

\label{subsec:dpo_vs_grpo_brief}

The recent success of group relative policy optimization~(GRPO)~\cite{guo2025deepseek} shows the effectiveness of outcome-based reinforcement learning for scaling complex reasoning, provided that reliable verifiers are available. GRPO primarily trains models through outcome-level rewards computed by rule-based or model-based verifiers~\cite{guo2025deepseek}. However, mitigating \textit{visual laziness} in multimodal models presents a different challenge. In the multimodal domain, a model can often produce a plausible, and sometimes even correct, response by relying on strong \textit{language priors} without truly anchoring its prediction in visual evidence~\cite{wang2025perception, long2026understanding, zhang2025mllms, zhou2024analyzing}.

Importantly, correctness is not a sufficient proxy for grounding. Prior work shows that models may achieve correct answers while relying on incorrect intermediate programs or inconsistent reasoning, essentially being ``right for the wrong reasons''~\cite{panagopoulou2025viunit, gandhi2022measuring, liang2025colorbench}. Outcome-only rewards can thus inadvertently reinforce \textit{causal misattribution}~\cite{niu2021counterfactual}, where the model reaches the right answer for purely linguistic reasons, preserving the very textual shortcuts we aim to remove.

Because our goal is to correct the model's insufficient reliance on visual evidence rather than merely improve the final response, \texttt{VIGIL} leverages the pairwise structure of DPO~\cite{rafailov2023direct} to directly compare the seeing state against a counterfactual blind state within a unified log-likelihood objective. This comparison explicitly suppresses \underline{blind confidence} by rewarding responses based on reliance on the visual input. In contrast, GRPO optimizes rewards over sampled responses in a group, and does not naturally encode such counterfactual state comparisons. Incorporating equivalent supervision would require constructing and evaluating additional blind trajectories for each sample, substantially increasing compute and memory. DPO therefore provides a highly efficient and stable offline optimization objective for decoupling the multimodal manifold from the textual manifold. More detailed discussions and empirical results on GRPO are provided in Appendix~\ref{sec:dpo_vs_grpo}.

\let\resizebox\oldresizebox
\makeatother
}
\section{Method}
\vspace{-4pt}
\label{sec:method}

To address \textit{visual laziness}, we propose \texttt{VIGIL}, a framework that explicitly anchors MLLM reasoning to visual evidence. It introduces a geometric constraint to maximize the mutual information between visual inputs and model responses, preventing the model from falling into language-driven optimization shortcuts.

\subsection{Grounding via Information Maximization}
\vspace{-4pt}
\label{subsec:theory}

We posit that an ideal MLLM should not merely generate high-probability tokens conditioned on \textit{language priors}, but should instead maximize the information dependency between the visual input $x_\text{v}$ and the generated response $y$.

\paragraph{Visual Information Gain (VIG).}
\vspace{-4pt}
To quantify this dependency, we draw inspiration from \textit{Maximum Mutual Information}~(MMI) principles~\cite{li2016diversity} and introduce VIG. We define VIG as the Pointwise Mutual Information~(PMI) between the visual input $x_\text{v}$ and the response $y$, conditioned on the instruction $x_\text{t}$:

\begin{definition}[Visual Information Gain]
For a given multimodal input tuple \normalfont{$(x_\text{v}, x_\text{t})$} and response \normalfont{$y$}, the Visual Information Gain is defined as:
\vspace{-4pt}
\begin{equation}
\label{eq:vig}
\normalfont{\text{VIG}(y, x_\text{v} | x_\text{t}) = \log \frac{\pi_\theta(y | x_\text{v}, x_\text{t})}{\pi_\theta(y | x^\emptyset_\text{v}, x_\text{t})}}
\vspace{-4pt}
\end{equation}
where $x^\emptyset_\text{v}$ denotes the counterfactual attention-masked blind state, functionally equivalent to performing a \textit{do-intervention} on the visual modality~\cite{niu2021counterfactual}.
\end{definition}

VIG measures the reduction in uncertainty about $y$ provided \textit{solely} by the visual evidence $x_\text{v}$. A VIG near zero indicates that the model is relying entirely on \textit{language priors}, a core symptom of visual laziness.

\begin{figure*}[!tb]
    \centering
    \includegraphics[width=\textwidth]{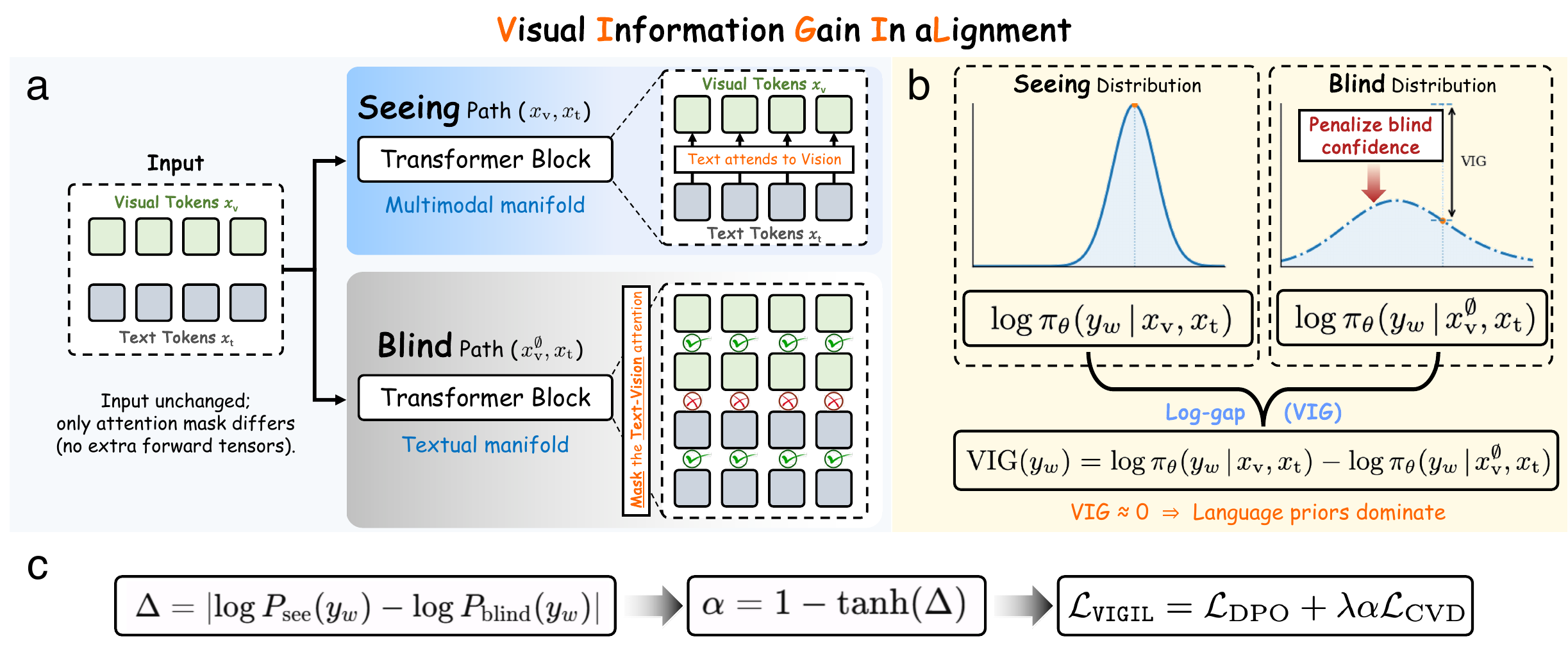}
    \caption{\textbf{Overview of \texttt{VIGIL}.} 
\textbf{(a)}~Dual-path forward pass. For each training sample, we run two matched conditions with identical image tokens, text tokens, positional embeddings, and projector outputs. In the \textbf{seeing} path, text tokens can attend to visual tokens normally. In the counterfactual \textbf{blind} path, the input tensors remain unchanged, but text-query rows are masked from visual-token key/value columns in every transformer layer, preventing text positions from accessing visual evidence through attention and yielding the blind state $x_\text{v}^{\emptyset}$. The two paths differ only in attention connectivity, without additional input construction or image corruption.
\textbf{(b)}~Geometric constraint via Visual Information Gain (VIG). We explicitly contrast the likelihood of the preferred response under seeing and blind states. \texttt{VIGIL} penalizes high \textbf{blind confidence} and enlarges the log-gap
$\mathrm{VIG}(y_w)=\log \pi_\theta(y_w\!\mid\!x_\text{v},x_\text{t})-\log \pi_\theta(y_w\!\mid\!x_\text{v}^{\emptyset},x_\text{t})$,
encouraging predictions to depend on visual evidence rather than language priors.
\textbf{(c)}~Dynamic gating and unified objective. A gating factor $\alpha$ is computed from the seeing--blind gap and adaptively balances the standard DPO loss with the Counterfactual Visual Decoupling (CVD) constraint, yielding $\mathcal{L}_\texttt{VIGIL}=\mathcal{L}_\text{DPO}+\lambda\alpha\mathcal{L}_\text{CVD}$.}    
\label{fig:method}
\end{figure*}

\paragraph{The Constrained Optimization Problem.}
\vspace{-4pt}
Standard DPO maximizes the margin between preferred ($y_w$) and disfavored ($y_l$) responses. However, as discussed in Sec.~\ref{sec:preliminaries}, maximizing the reward alone does not guarantee grounding. We instead formulate multimodal alignment as a \textit{constrained optimization problem}, where we maximize the preference reward subject to the constraint that the \textbf{Conditional Mutual Information} $I(y; x_\text{v} | x_\text{t})$ exceeds a threshold $\delta$:
\vspace{-4pt}
\begin{equation}
\label{eq:constrained_opt}
\max_{\pi_\theta} \quad \mathbb{E}_{\mathcal{D}} \left[ \mathcal{J}_{\text{DPO}}(\pi_\theta) \right]
\quad \text{s.t.} \quad \underbrace{\mathbb{E}_{\mathcal{D}} \left[ \text{VIG}(y_w, x_\text{v} | x_\text{t}) \right]}_{\approx I(y_w; x_\text{v} | x_\text{t})} \ge \delta
\vspace{-4pt}
\end{equation}

This constraint encourages the generated response $y_w$ contains sufficient information derived from $x_\text{v}$, thereby decoupling the multimodal manifold from the pure \textit{textual manifold} and preventing optimization from collapsing onto language-only shortcuts.

\paragraph{Derivation of the Counterfactual Visual Decoupling (CVD) Objective.}
\vspace{-4pt}
To solve Eq.~\ref{eq:constrained_opt}, we introduce a Lagrange multiplier $\lambda \ge 0$, leading to the following objective:
\vspace{-4pt}
\begin{equation}
\label{eq:lagrangian}
\mathcal{L}(\pi_\theta, \lambda) = \mathcal{J}_{\text{DPO}}(\pi_\theta) + \lambda \mathbb{E} \left[ \log \pi_\theta(y_w | x_\text{v}, x_\text{t}) - \log \pi_\theta(y_w | x^\emptyset_\text{v}, x_\text{t}) \right]
\end{equation}
We instantiate this Lagrangian term as a preference optimization problem, where the seeing state $(x_\text{v}, x_\text{t})$ is preferred over the attention-masked blind state $(x^\emptyset_\text{v}, x_\text{t})$ for the same response $y_w$. Applying the Bradley-Terry model, we derive our Counterfactual Visual Decoupling (CVD) loss:
\vspace{-4pt}
\begin{equation}
\label{eq:cvd_loss_final}
\mathcal{L}_{\text{CVD}}(\pi_\theta) = -\mathbb{E}_{\mathcal{D}} \left[ \log \sigma \left( \beta \log \frac{\pi_\theta(y_w | x_\text{v}, x_\text{t})}{\pi_{\text{ref}}(y_w | x_\text{v}, x_\text{t})} - \beta \log \frac{\pi_\theta(y_w | x^\emptyset_\text{v}, x_\text{t})}{\pi_{\text{ref}}(y_w | x^\emptyset_\text{v}, x_\text{t})} \right) \right]
\vspace{-4pt}
\end{equation}
This loss enforces a preference for the seeing state over the blind state by penalizing responses whose likelihood does not decrease when visual evidence is removed through counterfactual masking, while the reference model $\pi_{\mathrm{ref}}$ serves to normalize language priors. Importantly, the frozen reference policy is evaluated under the same attention masks as the policy model: the seeing term uses the seeing mask and the blind term uses the blind mask. Thus, Eq.~\eqref{eq:cvd_loss_final} compares matched seeing and blind states, rather than normalizing a blind policy term with a seeing reference term.

\subsection{\texttt{VIGIL} Implementation}
\label{subsec:cvd_imp}

The \texttt{VIGIL} framework is implemented through three integrated steps:

\paragraph{Step 1: Efficient Visual Counterfactuals.}
\vspace{-4pt}
To simulate the blind state $x^\emptyset_\text{v}$, we perform \textit{attention masking}. We zero out the attention weights between textual and visual tokens in the forward pass, ensuring the probability drop is strictly causal to the missing visual modality. Pseudocode is provided in Appendix~\ref{sec:blind_state}.

\paragraph{Step 2: The Geometric Decoupling Loss.}
\vspace{-4pt}
We utilize $\mathcal{L}_{\text{CVD}}$ (Eq.~\ref{eq:cvd_loss_final}) as a geometric constraint. This pushes the optimization trajectory to differentiate between grounded reasoning and prior-driven hallucination.

\paragraph{Step 3: Self-Adversarial Hard Negative Mining.}
\vspace{-4pt}
To complement our geometric constraints, we generate \textit{hallucination-like} negatives $y_{\text{hard}}$ by prompting the reference model to include plausible but absent details~\cite{qiu2025dadpo}. These samples provide sharper gradients for fine-grained visual discrimination.

\subsection{Unified Objective with Dynamic Gating}
\vspace{-4pt}
\label{subsec:unified}

To balance semantic alignment and visual grounding, we introduce a dynamic gating factor $\alpha$:
\vspace{-4pt}
\begin{equation}
\alpha = 1 - \tanh( | \log \pi_\theta(y_w | x_\text{v}, x_\text{t}) - \log \pi_\theta(y_w | x^\emptyset_\text{v}, x_\text{t}) | )
\vspace{-4pt}
\end{equation}
Here, $\alpha$ automatically reduces the regularization strength for samples that already exhibit strong visual grounding (i.e., a high VIG gap). The final \texttt{VIGIL} objective is defined as:
\vspace{-4pt}
\begin{equation}
\label{eq:total}
\mathcal{L}_\texttt{VIGIL} = \mathcal{L}_\text{DPO}(y_w, y_{\text{hard}}) + \lambda \cdot \alpha \cdot \mathcal{L}_\text{CVD}
\vspace{-4pt}
\end{equation}
This unified objective ensures the model learns not just what to say, but why to say it based on visual evidence.

\subsection{Facing the Sword of Damocles: The Fragile Equilibrium of Grounded Reasoning}
\vspace{-4pt}
\label{subsec:damocles}

The heavy reliance on \textit{language priors} in Multimodal Large Language Models is often described as a double-edged sword, or more aptly, a ``Sword of Damocles'' hanging over the model's reliability. While these priors enable models to generate fluent and contextually plausible narratives~\cite{chowdhery2023palm}, they exist as a fragile equilibrium that easily collapses when the reasoning task demands strict adherence to visual evidence rather than statistical correlation~\cite{zhang2025mllms, zhou2024analyzing}.  Our \texttt{VIGIL} framework acknowledges this inherent tension by moving beyond simple error correction. By introducing the counterfactual blind state $x^\emptyset_\text{v}$, we establish a regulatory mechanism that quantifies the risk of over-reliance on linguistic context. This creates what we term a \textbf{Geometric Information Bottleneck}, which ensures that the model's policy $\pi_\theta$ is causally anchored in the pixel space. Unlike existing reweighting schemes that treat all samples with equal priority~\cite{hu2026seeing}, \texttt{VIGIL} explicitly penalizes high confidence in the absence of visual support, transforming the passive vulnerability of priors into a controlled and grounded reasoning process. To sum up, the primary advantages of our approach are threefold. (1)~It offers a \textbf{geometric intervention} that is more fundamental than numerical loss reweighting, providing a direct path for causal alignment. (2)~It achieves \textbf{remarkable efficiency}, requiring neither additional reference models nor full-dataset fine-tuning to achieve state-of-the-art performance. (3)~By stabilizing the balance between linguistic fluency and visual fidelity, \texttt{VIGIL} encourages the emergence of spatial grounding capabilities, suggesting that teaching a model to avoid hallucinations inherently deepens its understanding of the visual world.
\section{Experiments}
\vspace{-4pt}
\label{sec:exp}

\subsection{Implementation Details}
\vspace{-4pt}
\label{subsec:impl}

We implement all alignment algorithms using the OpenRLHF framework~\cite{hu2025openrlhf}. Training is conducted on a high-performance computing cluster equipped with NVIDIA H100 (80GB) GPU to ensure efficient distributed execution. To manage the memory requirements of training a 72B parameter model with DPO, which necessitates maintaining both the policy and reference models simultaneously, we employ DeepSpeed ZeRO-3~\cite{rajbhandari2020zero} with CPU offloading for optimizer states. We also integrate FlashAttention-2~\cite{dao2024flashattention} to accelerate the processing of long-context multimodal sequences. For experiments at the 7B scale, including Qwen2.5-VL-7B~\cite{bai2023qwen} and LLaVA-OneVision-7B~\cite{li2024llava}, we utilize standard Fully Sharded Data Parallel (FSDP~\cite{zhao2023pytorch} strategies on A100 nodes. We perform full parameter fine-tuning for all models to maximize learning plasticity. In alignment with recent findings in multimodal preference optimization~\cite{qiu2025dadpo}, we use large global batch sizes to maintain training stability: 2048 for 72B models and 1024 for 7B models. All models are trained for exactly one epoch to prevent overfitting to the preference set. The learning rate is set to $5 \times 10^{-7}$ using a cosine decay scheduler with a warmup ratio of 0.03. The KL penalty coefficient $\beta$ is fixed at 0.1 across all DPO variants. For our proposed \texttt{VIGIL}, the dynamic gating factor $\alpha$ is computed on a per-batch basis. We implement counterfactual visual inputs efficiently by manipulating the attention mask during the forward pass rather than modifying physical input tensors, ensuring the blind forward pass incurs minimal computational overhead. To ensure a rigorous comparison, we reproduce all baselines, including SimPO~\cite{meng2024simpo}, HA-DPO~\cite{zhao2025beyond}, and DA-DPO~\cite{qiu2025dadpo}, on the same training data using the hyperparameters recommended in their official implementations.

\subsection{Evaluation Benchmarks}
\vspace{-4pt}
\label{subsec:benchmarks}

We adopt a multi-layered evaluation protocol to assess the trade-off between visual faithfulness and reasoning intelligence. For hallucination evaluation, we report F1-scores on the POPE benchmark~\cite{li2023evaluating} across its adversarial, popular, and random splits. As performance on POPE has begun to saturate for state-of-the-art models, we also prioritize AMBER~\cite{wang2024amber}, a generative benchmark requiring free-form descriptions verified against fine-grained annotations. This metric is sensitive to hallucination propagation issues in long-context generation. Following the protocol of DA-DPO~\cite{qiu2025dadpo}, we include MMHal-Bench~\cite{sun2024mmhal} to measure hallucination in realistic, open-ended VQA scenarios. 

Meanwhile, to verify that our visual constraints do not degrade general intelligence, we evaluate models on multimodal reasoning benchmarks MathVista~\cite{lu2024mathvista} and MMBench~\cite{liu2024mmbench}. Additionally, we perform evaluations on text-only benchmarks, namely MMLU~\cite{MMLU} and GSM8K~\cite{GSM8K}, to show that \texttt{VIGIL} improves multimodal reasoning without compromising text-only capabilities (Appendix~\ref{sec:capability_tax}).


\vspace{-4pt}
\subsection{Baselines}
\vspace{-4pt}
\label{subsec:baselines}

To demonstrate the effectiveness of \texttt{VIGIL}, we benchmark against a diverse array of alignment strategies categorized as follows:

\paragraph{General Alignment Methods.}
\vspace{-4pt}
We consider SFT as the performance baseline. For preference alignment, we employ standard DPO~\cite{rafailov2023direct} and SimPO~\cite{meng2024simpo}. SimPO is a reference-free algorithm that allows us to determine if performance gains stem from our visual constraints or simply from improved optimization stability. Comparing \texttt{VIGIL} against SimPO helps identify whether a superior optimizer is sufficient for multimodal grounding or if explicit visual modeling is required.

\paragraph{Hallucination-Specific Optimization.}
\vspace{-4pt}
We compare \texttt{VIGIL} against HA-DPO~\cite{zhao2025beyond}, which mitigates hallucination by training on synthetic negative captions generated through text rewriting. We also include the state-of-the-art DA-DPO~\cite{qiu2025dadpo}, which addresses overfitting by reweighting preference pairs based on difficulty. Unlike \texttt{VIGIL}, which utilizes physical interventions through counterfactual visual inputs, DA-DPO relies on the implicit difficulty distribution of the dataset.

\paragraph{Inference-Time Intervention.}
\vspace{-4pt}
To verify the advantages of training-time alignment over inference-time patching, we compare our method with visual contrastive decoding~(VCD)~\cite{he2025mitigating}. VCD penalizes hallucination by contrasting logits from original and distorted visual inputs during decoding. While effective, VCD increases inference latency and is sensitive to hyperparameter settings (See Appendix~\ref{sec:vcd}). \texttt{VIGIL} instead exposes the model to analogous visual contrast during post-training, resulting in an inference-efficient policy.

\vspace{-4pt}
\subsection{Main Results}
\vspace{-4pt}
\label{subsec:main_results}

\paragraph{Model Size Scaling Behavior.}
\vspace{-4pt}
We first assess the performance of \texttt{VIGIL} and competing methods as we scale the Qwen2.5-VL base model from 7B to 72B~(Table~\ref{tab:main_scaling}). \texttt{VIGIL} achieves state-of-the-art results across all evaluated metrics. A notable insight from these results is the scaling behavior of the model: while \texttt{VIGIL} improves the 7B model by 4.1 percentage points on POPE$_\text{Adv}$, this gain increases to 5.3 percentage points for the 72B model. This performance supports the existence of a post-training scaling law, suggesting that larger models are better positioned to interpret the visual counterfactuals introduced by our method. Whether this advantage arises from a higher knowledge capacity~\cite{allen2025physics} or from a more favorable latent representation geometry~\cite{liu2026dispersion} remains an interesting question for future study. In contrast, text-centric methods like SimPO show diminishing returns on multimodal tasks, indicating that optimization stability alone cannot compensate for the absence of modality-specific constraints.

\paragraph{Cross-Architecture Universality.}
\vspace{-4pt}
We further evaluate the universality of our approach on different base model architectures and designs~(Table~\ref{tab:generalization}). \texttt{VIGIL} consistently outperforms standard DPO across different architectures, including the projector-based LLaVA-OneVision-7B and the visual-centric InternVL2.5-26B~\cite{chen2024internvl}, which utilizes a 6B vision encoder. For both architectures, \texttt{VIGIL} achieves an improvement of approximately 4.0 point on POPE$_\text{Adv}$. These results demonstrate that even models with powerful visual encoders remain susceptible to visual laziness, a failure mode that our counterfactual masking effectively addresses.

\begin{table*}[!tb]
\caption{\textbf{Main Results across Model Scales.} We report performance on both Qwen2.5-VL-7B and 72B to demonstrate the scalability of \texttt{VIGIL}. Performance gains are more pronounced on the 72B model, supporting the hypothesis that stronger foundation models utilize counterfactual visual signals more effectively.}
\label{tab:main_scaling}
\vspace{-8pt}
\centering
\setlength{\tabcolsep}{3pt} 
\renewcommand{\arraystretch}{1.15} 

\resizebox{\textwidth}{!}{%
    \begin{tabular}{ll ccccc ccc}
    \toprule
    \multirow{2}{*}{\textbf{Base Model}} & \multirow{2}{*}{\textbf{Method}} & \multicolumn{5}{c}{\textbf{Hallucination}} & \multicolumn{2}{c}{\textbf{System-2 Reasoning}} & \multirow{2}{*}{\textbf{Avg Score}$\uparrow$}\\
    
    \cmidrule(lr){3-7} \cmidrule(lr){8-9}
    & & \texttt{POPE}$_\text{Rand}$~$\uparrow$
    & \texttt{POPE}$_\text{Pop}$~$\uparrow$
    & \texttt{POPE}$_\text{Adv}$~$\uparrow$
    & \texttt{AMBER}$_\text{Gen}$~$\uparrow$
    & \texttt{MMHal}$_\text{Score}$~$\uparrow$
    & \texttt{MathVista}~$\uparrow$
    & \texttt{MMBench}~$\uparrow$ \\

    \midrule
    \multirow{5}{*}{Qwen2.5-VL-\textbf{7B}~\cite{bai2023qwen}} & 
    \textcolor{gray}{SFT only}
    & \textcolor{gray}{85.1} & \textcolor{gray}{83.0} & \textcolor{gray}{80.5} & \textcolor{gray}{41.2} & \textcolor{gray}{34.5} & \textcolor{gray}{48.2} & \textcolor{gray}{70.5} & \textcolor{gray}{63.3} \\

    & \textcolor{gray}{SFT +} DPO~\cite{rafailov2023direct}
    & 86.2 & 84.5 & 82.8 & 42.5 & 36.8 & 48.0 & 71.2 & 64.6 \\
    
    & \textcolor{gray}{SFT +} SimPO~\cite{meng2024simpo}
    & 86.5~$_{\textcolor{forestgreen}{(+0.3)}}$ & 84.8~$_{\textcolor{forestgreen}{(+0.3)}}$ & 83.1~$_{\textcolor{forestgreen}{(+0.3)}}$ & 42.8~$_{\textcolor{forestgreen}{(+0.3)}}$ & 37.0~$_{\textcolor{forestgreen}{(+0.2)}}$ & 49.1~$_{\textcolor{forestgreen}{(+1.1)}}$ & 71.5~$_{\textcolor{forestgreen}{(+0.3)}}$ & 64.9~$_{\textcolor{forestgreen}{(+0.3)}}$ \\

    & \textcolor{gray}{SFT +} DA-DPO~\cite{qiu2025dadpo}
    & 87.0~$_{\textcolor{forestgreen}{(+0.8)}}$ & 85.8~$_{\textcolor{forestgreen}{(+1.3)}}$ & 84.2~$_{\textcolor{forestgreen}{(+1.4)}}$ & 44.8~$_{\textcolor{forestgreen}{(+2.3)}}$ & 38.5~$_{\textcolor{forestgreen}{(+1.7)}}$ & 48.8~$_{\textcolor{forestgreen}{(+0.8)}}$ & 72.0~$_{\textcolor{forestgreen}{(+0.8)}}$ & 65.9~$_{\textcolor{forestgreen}{(+1.3)}}$ \\

    & \cellcolor{lightblue!60}\textcolor{gray}{SFT +} \textbf{\texttt{VIGIL} (Ours)}
    & \cellcolor{lightblue!60}\textbf{88.5}~$_{\textcolor{forestgreen}{(+2.3)}}$ 
    & \cellcolor{lightblue!60}\textbf{87.2}~$_{\textcolor{forestgreen}{(+2.7)}}$ 
    & \cellcolor{lightblue!60}\textbf{86.9}~$_{\textcolor{forestgreen}{(+4.1)}}$ 
    & \cellcolor{lightblue!60}\textbf{46.5}~$_{\textcolor{forestgreen}{(+4.0)}}$ 
    & \cellcolor{lightblue!60}\textbf{40.2}~$_{\textcolor{forestgreen}{(+3.4)}}$ 
    & \cellcolor{lightblue!60}\textbf{49.5}~$_{\textcolor{forestgreen}{(+1.5)}}$ 
    & \cellcolor{lightblue!60}\textbf{72.5}~$_{\textcolor{forestgreen}{(+1.3)}}$ 
    & \cellcolor{lightblue!60}\textbf{67.3}~$_{\textcolor{forestgreen}{(+2.7)}}$ \\ 

    \midrule[\heavyrulewidth] 

    \multirow{6}{*}{Qwen2.5-VL-\textbf{72B}~\cite{bai2023qwen}} & \textcolor{gray}{SFT only} 
    & \textcolor{gray}{86.5} & \textcolor{gray}{84.2} & \textcolor{gray}{82.1} & \textcolor{gray}{45.3} & \textcolor{gray}{38.2} & \textcolor{gray}{54.5} & \textcolor{gray}{76.8} & \textcolor{gray}{66.8} \\

    & \textcolor{gray}{SFT +} DPO~\cite{rafailov2023direct}
    & 87.8 & 85.9 & 84.5 & 46.8 & 40.5 & 54.1 & 77.2 & 68.1 \\
    
    & \textcolor{gray}{SFT +} SimPO~\cite{meng2024simpo}
    & 88.1~$_{\textcolor{forestgreen}{(+0.3)}}$ & 86.2~$_{\textcolor{forestgreen}{(+0.3)}}$ & 84.8~$_{\textcolor{forestgreen}{(+0.3)}}$ & 47.0~$_{\textcolor{forestgreen}{(+0.2)}}$ & 41.1~$_{\textcolor{forestgreen}{(+0.6)}}$ & 55.2~$_{\textcolor{forestgreen}{(+1.1)}}$ & 77.5~$_{\textcolor{forestgreen}{(+0.3)}}$ & 68.6~$_{\textcolor{forestgreen}{(+0.5)}}$ \\

    & \textcolor{gray}{SFT +} HA-DPO~\cite{zhao2025beyond}
    & 88.5~$_{\textcolor{forestgreen}{(+0.7)}}$ & 87.1~$_{\textcolor{forestgreen}{(+1.2)}}$ & 86.2~$_{\textcolor{forestgreen}{(+1.7)}}$ & 48.5~$_{\textcolor{forestgreen}{(+1.7)}}$ & 42.8~$_{\textcolor{forestgreen}{(+2.3)}}$ & 53.8~$_{\textcolor{crimson}{(-0.3)}}$ & 76.5~$_{\textcolor{crimson}{(-0.7)}}$ & 69.1~$_{\textcolor{forestgreen}{(+1.0)}}$ \\

    & \textcolor{gray}{SFT +} DA-DPO~\cite{qiu2025dadpo}
    & 89.2~$_{\textcolor{forestgreen}{(+1.4)}}$ & 88.0~$_{\textcolor{forestgreen}{(+2.1)}}$ & 87.4~$_{\textcolor{forestgreen}{(+2.9)}}$ & 49.8~$_{\textcolor{forestgreen}{(+3.0)}}$ & 44.2~$_{\textcolor{forestgreen}{(+3.7)}}$ & 55.4~$_{\textcolor{forestgreen}{(+1.3)}}$ & 77.8~$_{\textcolor{forestgreen}{(+0.6)}}$ & 70.3~$_{\textcolor{forestgreen}{(+2.2)}}$ \\

    & \cellcolor{lightblue!60}\textcolor{gray}{SFT +} \textbf{\texttt{VIGIL} (Ours)}
    & \cellcolor{lightblue!60}\textbf{91.5}~$_{\textcolor{forestgreen}{(+3.7)}}$ 
    & \cellcolor{lightblue!60}\textbf{90.2}~$_{\textcolor{forestgreen}{(+4.3)}}$ 
    & \cellcolor{lightblue!60}\textbf{89.8}~$_{\textcolor{forestgreen}{(+5.3)}}$ 
    & \cellcolor{lightblue!60}\textbf{52.2}~$_{\textcolor{forestgreen}{(+5.4)}}$ 
    & \cellcolor{lightblue!60}\textbf{46.0}~$_{\textcolor{forestgreen}{(+5.5)}}$ 
    & \cellcolor{lightblue!60}\textbf{56.6}~$_{\textcolor{forestgreen}{(+2.5)}}$ 
    & \cellcolor{lightblue!60}\textbf{78.5}~$_{\textcolor{forestgreen}{(+1.3)}}$ 
    & \cellcolor{lightblue!60}\textbf{72.1}~$_{\textcolor{forestgreen}{(+4.0)}}$ \\ 

    \bottomrule
    \end{tabular}
}
\vspace{-8pt}
\end{table*}

\begin{table*}[!tb]
\caption{\textbf{Generalization across base model architectures.} We validate \texttt{VIGIL} across diverse architectures: LLaVA-OneVision and InternVL2.5-26B. \texttt{VIGIL} consistently outperforms standard DPO and DA-DPO. Performance gains are calculated relative to the standard DPO baseline. For the MMHal benchmark, we report the hallucination rate to provide a fine-grained measure of grounding failure.}
\label{tab:generalization}
\vspace{-8pt}
\centering
\setlength{\tabcolsep}{4.5pt}
\renewcommand{\arraystretch}{1.2}

\resizebox{\textwidth}{!}{%
    \begin{tabular}{l l ccc  ccc}
    \toprule
    \multirow{2}{*}{\textbf{Base Model}} & \multirow{2}{*}{\textbf{Method}} & \multicolumn{3}{c}{\textbf{Hallucination}} & \multicolumn{3}{c}{\textbf{General Capabilities}} \\
    \cmidrule(lr){3-5} \cmidrule(lr){6-8}
     & & \texttt{POPE}$_\text{Adv}$~$\uparrow$ & \texttt{AMBER}$_\text{Hal}$~$\downarrow$ & \texttt{MMHal}$_\text{HalRate}$~$\downarrow$ & \texttt{MathVista}~$\uparrow$ & \texttt{MMBench}~$\uparrow$ & \texttt{SeedBench}~$\uparrow$ \\
    \midrule
    
    \multirow{4}{*}{\shortstack[l]{\textbf{LLaVA-OneVision-7B}~\cite{li2024llava}\\ \textit{(SigLIP + MLP)}}} 
    & \textcolor{gray}{SFT only} 
    & \textcolor{gray}{80.5} & \textcolor{gray}{35.4} & \textcolor{gray}{2.76} & \textcolor{gray}{51.2} & \textcolor{gray}{72.1} & \textcolor{gray}{71.5} \\
    
    & \textcolor{gray}{SFT +} DPO~\cite{rafailov2023direct} 
    & 82.8 & 34.3 & 2.61 & 50.8 & 72.5 & 71.8 \\
    
    & \textcolor{gray}{SFT +}  DA-DPO~\cite{qiu2025dadpo} 
    & 84.2~$_{\textcolor{forestgreen}{(+1.4)}}$ & 28.0~$_{\textcolor{forestgreen}{(-6.3)}}$ & 2.78~$_{\textcolor{crimson}{(+0.17)}}$ & 51.5~$_{\textcolor{forestgreen}{(+0.7)}}$ & 73.0~$_{\textcolor{forestgreen}{(+0.5)}}$ & 72.4~$_{\textcolor{forestgreen}{(+0.6)}}$ \\
    
    & \cellcolor{lightblue!60}\textcolor{gray}{SFT +} \texttt{VIGIL} \textbf{(Ours)} 
    & \cellcolor{lightblue!60}\textbf{86.9}~$_{\textcolor{forestgreen}{(+4.1)}}$ 
    & \cellcolor{lightblue!60}\textbf{25.1}~$_{\textcolor{forestgreen}{(-9.2)}}$ 
    & \cellcolor{lightblue!60}\textbf{2.45}~$_{\textcolor{forestgreen}{(-0.16)}}$ 
    & \cellcolor{lightblue!60}\textbf{52.8}~$_{\textcolor{forestgreen}{(+2.0)}}$ 
    & \cellcolor{lightblue!60}\textbf{73.8}~$_{\textcolor{forestgreen}{(+1.3)}}$ 
    & \cellcolor{lightblue!60}\textbf{73.1}~$_{\textcolor{forestgreen}{(+1.3)}}$ \\
    \midrule
    
    \multirow{4}{*}{\shortstack[l]{\textbf{InternVL2.5-26B}~\cite{chen2024internvl}\\ \textit{(Large ViT Encoder)}}}
    & \textcolor{gray}{SFT only} 
    & \textcolor{gray}{84.1} & \textcolor{gray}{31.2} & \textcolor{gray}{2.50} & \textcolor{gray}{58.4} & \textcolor{gray}{79.2} & \textcolor{gray}{76.4} \\
    
    & \textcolor{gray}{SFT +} DPO~\cite{rafailov2023direct}
    & 85.5 & 29.8 & 2.35 & 57.9 & 79.5 & 76.8 \\
    
    & \textcolor{gray}{SFT +} DA-DPO~\cite{qiu2025dadpo}
    & 86.8~$_{\textcolor{forestgreen}{(+1.3)}}$ & 25.5~$_{\textcolor{forestgreen}{(-4.3)}}$ & 2.40~$_{\textcolor{crimson}{(+0.05)}}$ & 58.8~$_{\textcolor{forestgreen}{(+0.9)}}$ & 80.1~$_{\textcolor{forestgreen}{(+0.6)}}$ & 77.2~$_{\textcolor{forestgreen}{(+0.4)}}$ \\
    
    & \cellcolor{lightblue!60}\textcolor{gray}{SFT +} \texttt{VIGIL} \textbf{(Ours)} 
    & \cellcolor{lightblue!60}\textbf{89.4}~$_{\textcolor{forestgreen}{(+3.9)}}$ 
    & \cellcolor{lightblue!60}\textbf{22.3}~$_{\textcolor{forestgreen}{(-7.5)}}$ 
    & \cellcolor{lightblue!60}\textbf{2.15}~$_{\textcolor{forestgreen}{(-0.20)}}$ 
    & \cellcolor{lightblue!60}\textbf{60.1}~$_{\textcolor{forestgreen}{(+2.2)}}$ 
    & \cellcolor{lightblue!60}\textbf{81.3}~$_{\textcolor{forestgreen}{(+1.8)}}$ 
    & \cellcolor{lightblue!60}\textbf{78.0}~$_{\textcolor{forestgreen}{(+1.2)}}$ \\
    \bottomrule
    \end{tabular}
}
\vspace{-8pt}
\end{table*}

\vspace{-4pt}
\subsection{Ablation Study}
\vspace{-4pt}
\label{subsec:ablation}

We conduct ablation studies on Qwen2.5-VL-7B to analyze the source of performance gains, focusing on component contributions, mechanism design, system robustness, efficiency, and a surprising study on emergent spatial grounding.

\paragraph{Impact of Key Components.}
\vspace{-4pt}
The contribution of each module is shown in Table~\ref{tab:ablation_component}. Starting from the DPO baseline, the addition of hard negatives improves the POPE score by 2.0 percentage points, confirming the importance of high-quality preference data. The most significant improvement, a gain of 3.2 percentage points, is driven by the \textbf{Visual Anchor} ($x^\emptyset_\text{v}$). This result demonstrates that geometric contrast, specifically the ability to distinguish between the seeing and blind states, is essential for mitigating hallucinations. Moreover, the removal of dynamic gating results in a minor performance decrease, validating its role in stabilizing gradient flow during optimization.

\begin{table*}[!tb]
    \centering
    \begin{minipage}{0.48\textwidth}
        \centering
        \caption{\textbf{Component ablation.} We analyze the contribution of each module on Qwen2.5-VL-7B. The \textit{Visual Anchor} ($x^\emptyset_\text{v}$) is the most critical component, providing the largest performance boost.}
        \label{tab:ablation_component}
        \vspace{-8pt}
        \setlength{\tabcolsep}{3.5pt}
        \renewcommand{\arraystretch}{1.2}
        \resizebox{\textwidth}{!}{%
            \begin{tabular}{l ccc}
            \toprule
            \textbf{Method} & \texttt{POPE}$_\text{Rand}$~$\uparrow$ & \texttt{AMBER}$_\text{Gen}$~$\uparrow$ & \texttt{MathVista}~$\uparrow$ \\
            \midrule
            \rowcolor{lightblue!60} \textcolor{gray}{SFT +} \textbf{\texttt{VIGIL} (Ours)} & \textbf{88.5} & \textbf{46.5} & \textbf{49.5} \\

            $-$ Visual Anchor ($x^\emptyset_\text{v}$) 
            & 85.3~$_{\textcolor{crimson}{(-3.2)}}$ & 43.1~$_{\textcolor{crimson}{(-3.4)}}$ & 49.0~$_{\textcolor{crimson}{(-0.5)}}$ \\
            
            $-$ Hard Negatives 
            & 86.5~$_{\textcolor{crimson}{(-2.0)}}$ & 44.2~$_{\textcolor{crimson}{(-2.3)}}$ & 48.8~$_{\textcolor{crimson}{(-0.7)}}$ \\
            
            $-$ Dynamic Gating ($\alpha$) 
            & 87.4~$_{\textcolor{crimson}{(-1.1)}}$ & 45.4~$_{\textcolor{crimson}{(-1.1)}}$ & 49.2~$_{\textcolor{crimson}{(-0.3)}}$ \\
            
            \midrule
            \textcolor{gray}{SFT +} DPO~\cite{rafailov2023direct}
            & 86.2 & 42.5 & 48.0 \\
            \bottomrule
            \end{tabular}
        }
    \end{minipage}
    \hfill
    \begin{minipage}{0.48\textwidth}
        \centering
        \caption{\textbf{Alignment strategy comparison.} Our explicit masking (Geometric Decoupling) strategy significantly outperforms implicit difficulty reweighting (DA-DPO) and filtering.}
        \label{tab:mechanism_compare}
        \vspace{-8pt}
        \setlength{\tabcolsep}{5pt}
        \renewcommand{\arraystretch}{1.2}
        \resizebox{\textwidth}{!}{%
            \begin{tabular}{l cc}
            \toprule
            \textbf{Strategy} & \textbf{Core Mechanism} & \texttt{POPE}$_\text{Rand}$~$\uparrow$ \\
            \midrule
            DPO~\cite{rafailov2023direct} & Equality & 82.8 \\
            Filtering (10\%) & Remove Easy & 84.3~$_{\textcolor{forestgreen}{(+1.5)}}$ \\
            DA-DPO~\cite{qiu2025dadpo} & Reweighting & 85.7~$_{\textcolor{forestgreen}{(+2.9)}}$ \\
            \midrule
            \rowcolor{lightblue!60} \textbf{CVD-Masking} & Physical Constraint & \textbf{88.5}~$_{\textcolor{forestgreen}{(+5.7)}}$ \\
            \bottomrule
            \end{tabular}
        }
    \end{minipage}
\vspace{-12pt}
\end{table*}

\begin{table*}[!tb]
    \centering
        \begin{minipage}{0.48\textwidth}
        \centering
        \caption{\textbf{Performance across Visual Dependency Buckets.} Instead of generic difficulty, we categorize samples by \textit{Visual Dependency Index (VDI)}. \texttt{VIGIL} achieves substantial gains on samples with high visual dependency, demonstrating it effectively fixes ``blind'' hallucinations where baselines fail. Performances are evaluated on \texttt{POPE}$_\text{Rand}$.}
        \label{tab:dependency_buckets}
        \vspace{-8pt}
        \setlength{\tabcolsep}{5pt}
        \renewcommand{\arraystretch}{1.2}
        \resizebox{\textwidth}{!}{%
        \begin{tabular}{l c c cc}
            \toprule
            \multirow{3}{*}{\textbf{Method}} & \multicolumn{3}{c}{\textbf{VDI Bucket}} \\
            \cmidrule(lr){2-4}
            & \textbf{Low} & \textbf{Medium} & \textbf{High} \\
            & \small{(Text-heavy)} & \small{(In between)} & \small{(Fine-grained)} \\
            \midrule
            
            \textcolor{gray}{SFT +} DPO~\cite{rafailov2023direct} & 88.5 & 84.1 & 78.2 \\
            \textcolor{gray}{SFT +} DA-DPO~\cite{qiu2025dadpo} & 89.0 & 86.5 & 82.4 \\
            \rowcolor{lightblue!60}\textcolor{gray}{SFT +} \textbf{\texttt{VIGIL}} (Ours) & \textbf{89.2} & \textbf{88.8} & \textbf{87.5} \\
            \bottomrule
        \end{tabular}
        }
    \end{minipage}
    \hfill
    \begin{minipage}{0.48\textwidth}
        \centering
        \caption{\textbf{Effect of \texttt{VIGIL} on localization.} We evaluate Zero-shot Referring Expression Comprehension on \texttt{RefCOCOg} (val). Standard DPO often degrades spatial awareness. In contrast, \texttt{VIGIL} significantly improves localization accuracy, showcasing that our method physically anchors text to image regions.}
        \label{tab:refcoco}
        \vspace{-8pt}
        \setlength{\tabcolsep}{3.5pt}
        \renewcommand{\arraystretch}{1.2}
        \resizebox{\textwidth}{!}{%
            \begin{tabular}{l c cc}
                \toprule
                \multirow{2}{*}{\textbf{Method}} & \textbf{Hallucination} & \textbf{Localization} \\
                \cmidrule(lr){2-2} \cmidrule(lr){3-3}
                 & \texttt{POPE}$_\text{Adv}$ $\uparrow$ & \texttt{RefCOCOg} Acc@0.5 $\uparrow$ \\
                \midrule
                \textcolor{gray}{SFT only} & \textcolor{gray}{80.5} & \textcolor{gray}{45.2} \\
                \textcolor{gray}{SFT +} DPO~\cite{rafailov2023direct} & 82.8 & 44.8~$_{\textcolor{crimson}{(-0.4)}}$ \\
                \textcolor{gray}{SFT +} DA-DPO~\cite{qiu2025dadpo} & 84.2 & 46.1~$_{\textcolor{forestgreen}{(+0.9)}}$ \\
                \rowcolor{lightblue!60} \textcolor{gray}{SFT +} \textbf{\texttt{VIGIL} (Ours)} & \textbf{86.9} & \textbf{49.5}~$_{\textcolor{forestgreen}{(+4.3)}}$ \\
                \bottomrule
            \end{tabular}
        }
    \end{minipage}
\vspace{-12pt}
\end{table*}

\begin{figure*}[!tb]
    \centering
    \includegraphics[width=\textwidth]{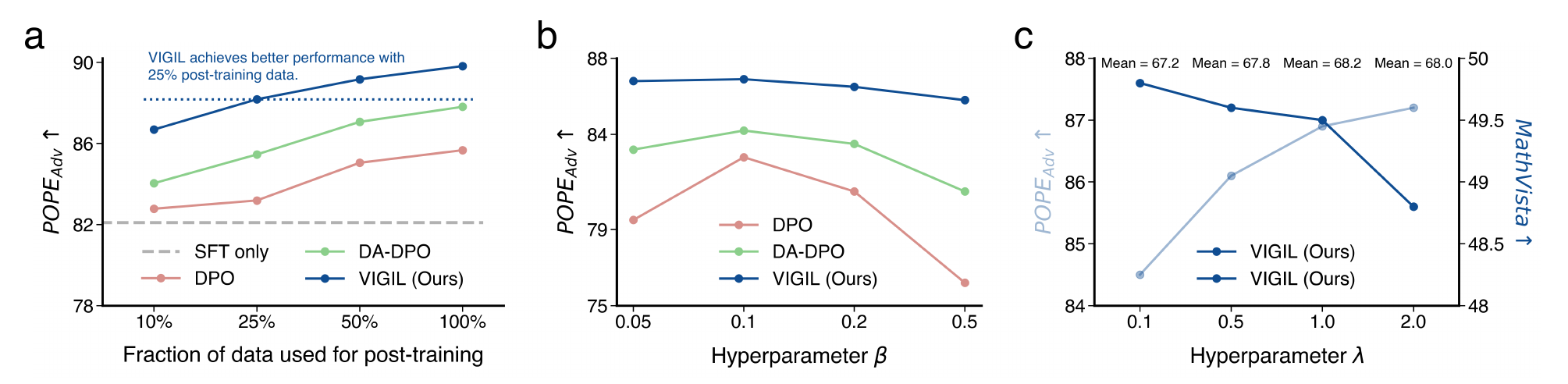}
    \vspace{-16pt}
    \caption{\textbf{Analyses of data efficiency and effects of hyperparameters.} \textbf{(a)} Our proposed \texttt{VIGIL} is data-efficient. It achieves performance comparable to state-of-the-art methods using only 25\% of the post-training data.
    \textbf{(b)} Our method is more robust to the KL penalty coefficient $\beta$ than competing methods, and the results justifies the choice of $\beta=0.1$. \textbf{(c)} The weighing coefficient $\lambda=1.0$ leads to a good balance between hallucination mitigation and reasoning performance.}
    \label{fig:ablation}
    \vspace{-8pt}
\end{figure*}

\paragraph{Mechanism Analyses.}
\vspace{-4pt}
We evaluate why \texttt{VIGIL} outperforms existing hard-negative mining techniques. Table~\ref{tab:mechanism_compare} compares our explicit masking strategy with the implicit reweighting approach used in DA-DPO~\cite{qiu2025dadpo}. While reweighting provides a 2.9 point gain, the physical masking employed by \texttt{VIGIL} yields a 5.3 point increase. This suggests that physical intervention on the input geometry provides a more effective supervision signal than numerical adjustments to loss weights. 

To further investigate this effect, Table~\ref{tab:dependency_buckets} categorizes samples by their Visual Dependency Index (VDI), which measures the divergence in model predictions when the visual input is removed. Standard DPO performs poorly on high-dependency samples, scoring 78.2, whereas \texttt{VIGIL} increases this to 87.5. These findings indicate that our method effectively guides the model to prioritize visual features when \textit{language priors} are ambiguous.

Additional rigorous evaluations of \texttt{VIGIL} on its optimization trajectory through the lens of \textit{Maximum Mutual Information}~(MMI)~\cite{li2016diversity} and \textit{Rate-Distortion Theory} are shown in Appendix~\ref{sec:info_theoretic_analysis}.

\paragraph{Robustness and Efficiency.}
\vspace{-4pt}
Figure~\ref{fig:ablation} illustrates the practical advantages of our framework. Regarding data efficiency, \texttt{VIGIL} trained on 25\% of the dataset matches the performance of DA-DPO trained on the full dataset. Despite the overhead of masking, this efficiency reduces the total training wall-clock time by approximately 70\% compared to full-data training (see Appendix~\ref{sec:efficiency}). In terms of stability, \texttt{VIGIL} remains robust across a range of KL penalty coefficients, $\beta \in [0.05, 0.5]$, whereas standard DPO often exhibits instability at lower $\beta$ values. We attribute this stability to the regularization effect of the visual anchor.

\paragraph{Emergent spatial grounding.}
\vspace{-4pt}
Table~\ref{tab:refcoco} shows that \texttt{VIGIL} improves zero-shot RefCOCOg performance by 4.3 percentage points. Although the model is not explicitly trained on bounding box coordinates, it develops improved spatial localization capabilities to satisfy the counterfactual grounding objective.

\vspace{-8pt}
\subsection{Qualitative Analysis}
\label{subsec:visualization}

\begin{figure*}[!tb]
    \centering
    \includegraphics[width=\textwidth]{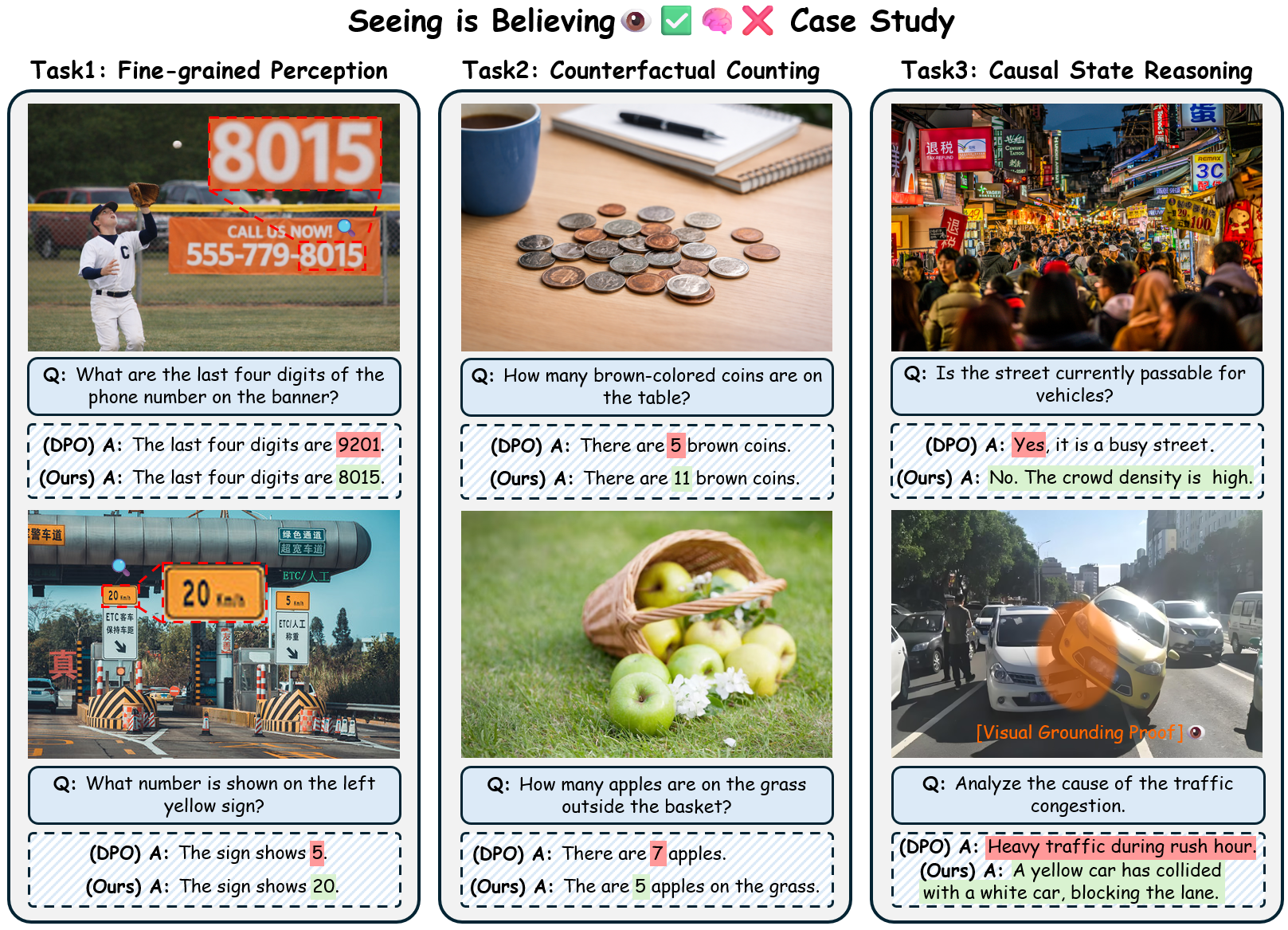} 
    \caption{\textbf{Qualitative Comparison of Visual Grounding.} 
    MLLMs often exhibit \textit{visual laziness} by relying on \textit{language priors}, resulting in plausible but factually incorrect hallucinations (red). In contrast, \texttt{VIGIL} anchors reasoning to visual evidence $x_\text{v}$ (green). 
    \textbf{Task 1 (Left):} In fine-grained perception, \texttt{VIGIL} accurately identifies the digits \underline{8015} as verified by the visual grounding box, whereas the DPO baseline hallucinates \underline{9201}. 
    \textbf{Task 2 (Middle):} In counterfactual counting, the DPO baseline defaults to generic numbers based on priors, while \texttt{VIGIL} correctly identifies \underline{11 coins} and \underline{5 apples}. 
    \textbf{Task 3 (Right):} In causal reasoning, instead of providing generic descriptions, \texttt{VIGIL} identifies specific physical states, such as vehicle collisions, to derive correct logical conclusions.}
    \label{fig:qualitative}
\vspace{-12pt}
\end{figure*}

To intuitively demonstrate how \texttt{VIGIL} mitigates \textit{visual laziness}, we present a qualitative comparison against the standard DPO baseline in Figure~\ref{fig:qualitative}. Both models are based on the Qwen2.5-VL-7B architecture. The results are categorized into three levels of difficulty: fine-grained perception, counterfactual counting, and causal state reasoning.

\paragraph{Fine-grained Perception.}
\vspace{-4pt}
The first case illustrates the model's ability to resolve high-frequency visual details. In the baseball scenario, when asked for the phone number on the banner, the DPO baseline predicts \underline{9201}, a response that appears plausible based on inherent \textit{language priors} but is factually incorrect. In contrast, \texttt{VIGIL} accurately identifies the digits \underline{8015}. The visual grounding proof, highlighted by the red bounding box, confirms that the model's attention is correctly anchored to the specific pixel region. This indicates that the model effectively utilizes visual evidence before generating the corresponding tokens.

\paragraph{Visual Counting.}
\vspace{-4pt}
Counting serves as a benchmark for visual grounding. As shown in the second case, the baseline exhibits a strong dependency on \textit{language priors}. For the brown coins and apples, the DPO baseline predicts \underline{5} and \underline{7}, respectively. These numbers are statistically frequent in training corpora yet incorrect for these specific images. By incorporating geometric constraints, \texttt{VIGIL} suppresses prior-based guessing and correctly identifies \underline{11 coins} and \underline{5 apples}. This demonstrates that our method encourages the model to perform actual enumeration rather than relying on probabilistic retrieval from its language components.

\paragraph{Causal State Reasoning.}
\vspace{-4pt}
The third case highlights the capability gap in reasoning about physical states. In the traffic accident scene, the baseline attributes the congestion to \underline{heavy traffic during rush hour}, representing a generic guess that ignores visual reality. Conversely, \texttt{VIGIL} identifies the specific causal mechanism, stating that a yellow car has collided with a white car. Similarly, in the crowded street scene, our model correctly identifies high crowd density as the physical barrier to vehicle passage. These observations confirm that \texttt{VIGIL} enables grounded reasoning, where logical conclusions are derived from accurate visual premises rather than language-based associations.

\vspace{-4pt}
\section{Conclusion}
\vspace{-4pt}
\label{sec:conclusion}

In this paper, we propose \texttt{VIGIL}, a principled post-training framework that mitigates hallucinations in MLLMs by maximizing visual information gain. Unlike standard preference optimization that often leads to \textit{visual laziness}, our approach introduces a geometric constraint through counterfactual visual decoupling. By penalizing model confidence in the blind state $x^\emptyset_\text{v}$, we force the optimization trajectory to anchor reasoning in actual visual features $x_\text{v}$ rather than inherent \textit{language priors}. Our evaluations demonstrate that \texttt{VIGIL} consistently outperforms state-of-the-art baselines while exhibiting superior scaling behavior and data efficiency. Notably, the emergence of spatial grounding capabilities suggests that our geometric anchoring effectively aligns the model's latent perception with its explicit generation. 

Despite its effectiveness, a potential limitation of our current framework lies in the simplistic construction of the counterfactual state via zero-out masking, which may not capture more nuanced cross-modal conflicts in highly cluttered environments. Future work could explore more granular counterfactual interventions, such as object-level spectral filtering or semantic-preserving transformations, to further refine the model's visual sensitivity. We hope our work encourages a shift toward causal and information-theoretic alignment in multimodal learning.

\makeatletter
\let\@vspace\old@vspace
\let\@vspacer\old@vspacer
\makeatother

\section*{Acknowledgements}
This research is supported in part by the U.S. National Science Foundation (OAC-2118240, HDR Institute: Imageomics).

\bibliographystyle{unsrtnat}
\bibliography{main}

\clearpage


\renewcommand{\thefigure}{S\arabic{figure}}
\renewcommand{\theHfigure}{S\arabic{figure}}
\setcounter{figure}{0}
\renewcommand{\thetable}{S\arabic{table}}
\renewcommand{\theHtable}{S\arabic{table}}
\setcounter{table}{0}

\beginappendix

\section{Related Work}
\label{sec:related}

\subsection{Mechanisms of Hallucination in MLLMs}
Recent Multimodal Large Language Models (MLLMs), such as Qwen-VL~\cite{bai2023qwen} and LLaVA-OneVision~\cite{li2024llava}, have achieved significant progress by connecting visual encoders with powerful LLMs. However, hallucination remains a persistent challenge where models generate content inconsistent with visual inputs~\cite{li2023evaluating, chen2025true, guo2025beyond, liu2025small, xu2025defacto, bai2024hallucination, peng2026group, wang2026visuallyguided}. Early hypotheses attributed this to the limited resolution of visual encoders or low-quality training data. However, recent mechanistic interpretability studies offer a different perspective~\cite{cao2025ground, zhu2026leveraging}. For instance, Zhang et al.~\cite{zhang2025mllms} reveal that visual information is often preserved in the model's latent space but is suppressed by strong \textit{language priors} during decoding. This suggests that the model effectively sees the correct object in its internal states but speaks the wrong answer due to learned language correlations. To address this, inference-time intervention methods like VCD~\cite{leng2024mitigating} and OPERA~\cite{huang2024opera} have been proposed. These methods suppress hallucinations by penalizing logits that rely heavily on \textit{language priors} during the decoding process. While effective, these inference-time methods increase latency and exhibit sensitivity to hyperparameter tuning. Our work seeks to internalize these constraints directly into the model weights during post-training, offering a more inference-efficient solution.

\subsection{Alignment and Efficiency in Multimodal Learning}
Reinforcement Learning from Human Feedback (RLHF)~\cite{ouyang2022training} and Direct Preference Optimization (DPO)~\cite{rafailov2023direct} are the standard methods for aligning models with human intent. In the multimodal domain, DPO variants like HA-DPO~\cite{zhao2025beyond} and DA-DPO~\cite{qiu2025dadpo} have been developed to reduce hallucinations. Multimodal visual generation has also explored cross-modal image representations and hybrid preference optimization for personalized video synthesis~\cite{zhao2024thinimg, Li_2025_ICCV}. DA-DPO, for instance, identifies difficulty imbalances in training data and uses an implicit reweighting strategy to emphasize harder samples.

Beyond static data reweighting, recent works explore dynamic efficiency. AdaptVision~\cite{lin2026adaptvision} utilizes reinforcement learning to dynamically select visual tokens to reduce computational overhead. Complementary efficiency work reduces training cost through generative dataset distillation~\cite{zhao2026hieramp} or task-aware low-rank adaptation~\cite{xiao2026not}. However, such methods either change the training data, change the adaptation parameterization, or rely on Online RL, which can introduce training instability and implementation complexity. In contrast, \texttt{VIGIL} operates within the stable Offline DPO framework. We achieve high data efficiency not by dynamic sampling, but by maximizing the information gain from each preference pair through geometric constraints.

\subsection{Architectural vs. Geometric Solutions}
To improve visual grounding and reasoning, one line of research focuses on architectural enhancements. Methods like VGent~\cite{kang2026vgent} adopt a modular design, employing external tools or separate reasoning modules to handle complex visual tasks. Recent benchmarks further show that dynamic spatial reasoning requires integrating perception with memory over time~\cite{liao2026spamem}. While modular designs can improve performance on specific benchmarks, they compromise the elegance and generalizability of end-to-end learning.

An alternative approach uses visual counterfactuals to enforce grounding without architectural changes. This concept, rooted in causal inference~\cite{pearl2009causality}, involves reasoning about how the model's prediction would change if the visual input were different. Previous works like ALBEF~\cite{li2021align} applied this in pre-training. Our work integrates this concept into preference learning. Unlike modular approaches that rely on external components, \texttt{VIGIL} introduces a geometric constraint by formulating the counterfactual \textit{blind state} ($x^\emptyset_\text{v}$) as a negative anchor. This forces the model to distinguish between seeing and blindness, thereby grounding preference optimization in visual evidence without requiring additional parameters.

\subsection{DPO vs. GRPO}
\label{sec:dpo_vs_grpo}

Group Relative Policy Optimization (GRPO)~\cite{guo2025deepseek} is widely recognized for its strengths in reinforcement learning fine-tuning. In particular, it provides a stable policy-gradient update by using group-relative advantages, and it scales well to long-horizon, multi-step reasoning when a reliable reward signal is available. These properties make GRPO a strong default choice for tasks with verifiable outcomes, such as program synthesis and math problems with deterministic checkers.

In \texttt{VIGIL}, however, our target is multimodal hallucination mitigation, where the primary failure mode is not incorrect reasoning per se, but incorrect \emph{modality dependence}. As analyzed in Sec.~\ref{sec:preliminaries}, an MLLM can produce a plausible or even correct final answer while still being driven by language priors rather than visual evidence. Under this problem setting, optimizing final-answer rewards alone does not reliably distinguish \emph{grounded} responses from \emph{prior-driven} responses, and may inadvertently reinforce visual laziness.

\paragraph{Why DPO fits our setting.}
DPO directly optimizes pairwise preferences and therefore does not require an external verifier to produce dense and reliable rewards. This is important for hallucination benchmarks, where correctness can be ambiguous and where the key objective is to align the \emph{dependency path} (using $x_v$) rather than only the terminal string. Moreover, DPO naturally supports controlled comparisons under matched conditions. This property is essential for our counterfactual design: we contrast the same instruction and candidate response under a seeing state $(x_v,x_t)$ and a blind state $(x_v^{\emptyset},x_t)$, and we penalize blind confidence while preserving the ability to answer correctly when vision is available. This yields a direct and low-variance training signal for grounding.

\paragraph{GRPO is computationally inefficient for our use case.}
To reproduce the same counterfactual logic in GRPO, one would need to construct multiple blind variants for each sample within a group and evaluate reward signals for each variant, since the key supervision comes from comparing seeing versus blind behavior. This substantially increases the number of forward passes, reward computations, and memory footprint. The overhead becomes especially pronounced for large-scale MLLMs, where GRPO already requires group rollouts and reward evaluation. In contrast, DPO allows us to integrate the seeing/blind contrast directly into a single preference-style objective, making the counterfactual constraint practical at scale.

\paragraph{Empirical comparison.}
We additionally implement a Visual-GRPO baseline on Qwen2.5-VL-7B, using POPE accuracy as the reward. As shown in Table~\ref{tab:grpo_compare}, Visual-GRPO is competitive on MathVista, which aligns with GRPO's known advantages on reasoning tasks. However, it underperforms on hallucination benchmarks. The results support our main point: for multimodal alignment, explicitly enforcing visual dependence via counterfactual geometric decoupling is more effective than reward-based policy optimization that primarily targets final-answer correctness.

\begin{table}[ht!]
\caption{\textbf{Comparison with GRPO.} Evaluations on Qwen2.5-VL-7B. \texttt{VIGIL} outperforms the GRPO-based reward-tuning in mitigating hallucinations while maintaining comparable reasoning capabilities.}
\label{tab:grpo_compare}
\setlength{\tabcolsep}{6pt}
\renewcommand{\arraystretch}{1.15} 
\centering
\resizebox{0.6\linewidth}{!}{%
\begin{tabular}{l ccc}
\toprule
\textbf{Method} & \texttt{POPE}$_\text{Adv}$~$\uparrow$ & \texttt{MMHal}~$\uparrow$ & \texttt{MathVista}~$\uparrow$ \\
\midrule
Visual-GRPO & 84.1 & 37.5 & \textbf{50.1} \\
\rowcolor{lightblue!60} \textbf{\texttt{VIGIL} (Ours)} & \textbf{86.9} & \textbf{40.2} & 49.5 \\
\bottomrule
\end{tabular}
}
\end{table}

\section{Blind State Construction}
\label{sec:blind_state}

To construct the counterfactual blind state $x_\mathrm{v}^{\emptyset}$, we keep the same image, visual tokens, projector outputs, text tokens, and positional embeddings as in the seeing path, and intervene only on attention connectivity. Specifically, a per-layer attention mask prevents text-query positions from attending to visual key/value positions, while leaving the vision encoder untouched. 

Table~\ref{tab:blind_state} compares this design with image-level perturbations such as blacking out, blurring, and shuffling. Attention masking achieves the best performance, whereas the black-image variant performs worst, suggesting that directly corrupting the image introduces undesirable distribution shifts. Thus, VIG is computed between matched seeing and blind states, which renders it a matched-state estimate rather than a corrupted-image likelihood ratio.

\begin{table}[ht!]
\caption{\textbf{Effect of different blind state construction techniques.} Compared with image-level perturbations, attention masking achieves the best performance while avoiding additional image processing cost.}
\label{tab:blind_state}
\setlength{\tabcolsep}{6pt}
\renewcommand{\arraystretch}{1.15} 
\centering
\resizebox{0.6\linewidth}{!}{%
\begin{tabular}{lcccc}
\toprule & Black & Blur & Shuffle & \cellcolor{lightblue!60} Attention Mask \\
\midrule
\texttt{POPE}$_\text{Adv}$~$\uparrow$ & 84.7 & 85.6 & 85.9 & \cellcolor{lightblue!60} \textbf{86.9} \\
\bottomrule
\end{tabular}
}
\end{table}

The pseudocode of blind state construction is provided below.

\begin{algorithm}[ht!]
\caption{Attention-Masked Blind State Construction for \texttt{VIGIL}}
\label{alg:blind_state}
\begin{algorithmic}[1]
\Require Image $x_\mathrm{v}$, instruction $x_\mathrm{t}$, response prefix $y_{<t}$, MLLM with $L$ language layers
\Require Visual-token index set $\mathcal{I}_\mathrm{v}$ after the vision projector; text-token index set $\mathcal{I}_\mathrm{t}$
\State Encode the real image once: $z_\mathrm{v}\leftarrow \mathrm{Proj}(\mathrm{Enc}_\mathrm{v}(x_\mathrm{v}))$
\State Build the shared token sequence $z\leftarrow[z_\mathrm{v};\mathrm{Emb}(x_\mathrm{t},y_{<t})]$ with the original positional embeddings
\State Construct the normal causal/padding attention mask $M_\mathrm{see}$
\State Initialize the blind mask $M_\mathrm{blind}\leftarrow M_\mathrm{see}$
\For{each language layer $\ell=1,\ldots,L$}
    \For{each text-query position $i\in\mathcal{I}_\mathrm{t}$}
        \For{each visual key/value position $j\in\mathcal{I}_\mathrm{v}$}
            \State Set $M_\mathrm{blind}^{(\ell)}[i,j]\leftarrow -\infty$
        \EndFor
    \EndFor
\EndFor
\State Run the seeing path with $(z,M_\mathrm{see})$ to obtain $\log\pi_\theta(y\mid x_\mathrm{v},x_\mathrm{t})$
\State Run the blind path with the same $z$ and $(M_\mathrm{blind})$ to obtain $\log\pi_\theta(y\mid x_\mathrm{v}^{\emptyset},x_\mathrm{t})$
\State \textbf{return} matched seeing and blind log-likelihoods for the CVD and VIG objectives
\end{algorithmic}
\end{algorithm}

The intervention in Algorithm~\ref{alg:blind_state} is applied after visual encoding and projection. Therefore, the blind path never feeds a corrupted image to the vision encoder and never changes the visual-token distribution. Only text-query rows are blocked from reading visual key/value columns. Visual tokens can still participate in their own preprocessing, and the causal, padding, and position masks remain identical to the seeing path. This matched construction ensures that the difference between the two likelihoods is attributable to cross-modal access rather than OOD image corruption.

\section{Preference Set, Reference Policy and Reproducibility}
\label{sec:reference_policy}

This section specifies the preference data, reference-policy evaluation, and judging protocol used for reproducibility.

\paragraph{Matched reference-policy evaluation.}
The frozen reference policy is evaluated under the same state as the policy model. In the seeing term of Eq.~\ref{eq:cvd_loss_final}, both $\pi_\theta$ and $\pi_\mathrm{ref}$ use $M_\mathrm{see}$. In the blind term, both models use $M_\mathrm{blind}$. Thus, the CVD objective compares matched seeing and blind states:
\begin{equation}
\label{eq:matched_ref_appendix}
\log \frac{\pi_\theta(y_w\mid x_\mathrm{v},x_\mathrm{t};M_\mathrm{see})}{\pi_\mathrm{ref}(y_w\mid x_\mathrm{v},x_\mathrm{t};M_\mathrm{see})}
-
\log \frac{\pi_\theta(y_w\mid x_\mathrm{v}^{\emptyset},x_\mathrm{t};M_\mathrm{blind})}{\pi_\mathrm{ref}(y_w\mid x_\mathrm{v}^{\emptyset},x_\mathrm{t};M_\mathrm{blind})}.
\end{equation}
This avoids normalizing a blind policy term with a seeing reference term and keeps the contrast focused on visual access rather than reference-model mismatch.

\paragraph{Preference data composition.}
The preference set contains 120K pairs. All methods are trained for one epoch from the same base checkpoint, and DPO, DA-DPO, and \texttt{VIGIL} use the same preference pool unless otherwise stated. Table~\ref{tab:pref_composition} reports the composition.

\begin{table}[t!]
\caption{\textbf{Preference-set composition.} The training pool combines hallucination-specific samples with general multimodal reasoning, OCR/chart, and math/spatial tasks.}
\label{tab:pref_composition}
\setlength{\tabcolsep}{6pt}
\renewcommand{\arraystretch}{1.15}
\centering
\resizebox{0.9\linewidth}{!}{%
\begin{tabular}{lcc}
\toprule
\textbf{Category} & \textbf{Fraction} & \textbf{Role in Training} \\
\midrule
Hallucination / grounding & 45\% & object faithfulness and visual premise correction \\
General VQA / reasoning & 25\% & broad multimodal instruction following \\
OCR / chart understanding & 15\% & text-rich and structured visual evidence \\
Math / spatial reasoning & 15\% & compositional and geometry-sensitive reasoning \\
\bottomrule
\end{tabular}
}
\end{table}

\paragraph{Visual CoT consistency judging.}
For VCC, GPT-4o receives the image, a ground-truth object list, and the first visual premise in the model's chain of thought. The judge returns a binary consistency label and, when inconsistent, the offending object or visual claim. The exact judging prompt used in our evaluation is:
\begin{quote}
\small
You are given an image, a list of ground-truth objects visible in the image, and the first visual premise extracted from a model's reasoning. Decide whether the premise is visually consistent with the image. Answer with \texttt{CONSISTENT} or \texttt{INCONSISTENT}. If inconsistent, name the offending object, attribute, count, or spatial relation. Do not judge later reasoning steps or the final answer; evaluate only the first visual premise against the image evidence.
\end{quote}

\paragraph{Manual audit.}
We manually audit 300 randomly sampled VCC judgments. The GPT-4o judge agrees with human annotation on 92.3\% of the audited samples. Most disagreements come from ambiguous attributes or partially occluded objects, rather than from systematic bias toward any training method. We use the automatic VCC judge only as an auxiliary grounding diagnostic and do not train on its labels.

\section{Maintenance of Text-Only Capabilities}
\label{sec:capability_tax}

To assess whether improved visual grounding comes at the expense of general reasoning ability, we evaluate \texttt{VIGIL} on text-only benchmarks with the image branch disabled, alongside representative multimodal reasoning benchmarks.

As shown in Table~\ref{tab:capability_tax}, \texttt{VIGIL} preserves text-only performance, with changes on MMLU and GSM8K limited to less than 0.2 points. Across three random seeds, the standard deviation on these benchmarks is approximately $\pm0.3$, indicating that the observed differences fall within run-to-run variation rather than systematic degradation. In contrast, \texttt{VIGIL} consistently improves multimodal reasoning performance on MathVista and MMBench. These results suggest that the CVD objective encourages visual dependence only when visual evidence is informative, rather than imposing an indiscriminate reliance on the visual modality. Consequently, \texttt{VIGIL} improves multimodal grounding without introducing a measurable capability tax or catastrophic forgetting of text-only reasoning skills.

\begin{table}[ht!]
\caption{\textbf{Text-only capability retention after \texttt{VIGIL} post-training.} \texttt{VIGIL} preserves text-only reasoning performance on MMLU and GSM8K while improving multimodal reasoning benchmarks. The small changes on text-only tasks fall within run-to-run variation, indicating no measurable capability tax.}
\label{tab:capability_tax}
\setlength{\tabcolsep}{6pt}
\renewcommand{\arraystretch}{1.15} 
\centering
\resizebox{0.8\linewidth}{!}{%
\begin{tabular}{lcccc}
\toprule
\multirow{2}{*}{\textbf{Method}} & \multicolumn{2}{c}{\textbf{Text-Only}} & \multicolumn{2}{c}{\textbf{Multimodal}} \\
\cmidrule(lr){2-3} \cmidrule(lr){4-5}
& \texttt{MMLU}~$\uparrow$ & \texttt{GSM8K}~$\uparrow$ & \texttt{MathVista}~$\uparrow$ & \texttt{MMBench}~$\uparrow$ \\
\midrule
\textcolor{gray}{SFT only} & 74.8 & 91.3 & 48.2 & 70.5 \\
\rowcolor{lightblue!60} \textcolor{gray}{SFT +} \texttt{VIGIL} \textbf{(Ours)} & 74.6 & 91.1 & \textbf{49.5} & \textbf{72.5} \\
\bottomrule
\end{tabular}
}
\end{table}

\section{Training Efficiency and Computational Resource Analysis}
\label{sec:efficiency}

We further evaluate the computational cost of \texttt{VIGIL} to demonstrate its practical feasibility for large-scale multimodal alignment. As shown in Table~\ref{tab:gpu_hours}, \texttt{VIGIL} exhibits a distinct advantage in resource utilization compared to standard DPO and its variants.

\paragraph{High Data Leverage.} 
The most significant efficiency gain stems from our method's high signal-to-noise ratio in causal anchoring. While standard DPO requires the full preference dataset to achieve convergence, \texttt{VIGIL} matches the performance of 100\% DA-DPO using only 25\% of the training samples. This reduction in data requirements directly translates to a \textbf{70\% decrease in total training wall-clock time}. For the Qwen2.5-VL-7B model, the entire alignment process is completed within 3.5 GPU hours on a single A100 node, representing a substantial improvement over existing benchmarks.

\paragraph{Minimal Forward Overhead.} 
Contrary to intuition, our dual-path (Seeing and Blind) optimization does not double the computational burden. Since the counterfactual blind state $x^\emptyset_\text{v}$ is implemented via \textbf{attention masking} rather than modifying physical input tensors, it bypasses the redundant re-processing of high-resolution visual tokens. The additional TFLOPs attributable to the CVD constraint account for less than 1\% of the total forward-backward pass.

\paragraph{Comparison with Online RL.} 
Compared to online reinforcement learning methods like GRPO or AdaptVision, which require massive sampling and multiple reference models during training, \texttt{VIGIL} operates within a stable offline framework. This avoids the memory overhead of maintaining group-relative samples and significantly reduces the VRAM requirement, allowing for the full-parameter fine-tuning of 72B-scale models on standard infrastructure without aggressive quantization.

\begin{table}[ht!]
\caption{\textbf{Computational Cost and Efficiency Comparison.} All models are evaluated on Qwen2.5-VL-7B. GPU hours are measured as the total time required to reach state-of-the-art performance (POPE $>$ 86.0).}
\label{tab:gpu_hours}
\setlength{\tabcolsep}{6pt}
\renewcommand{\arraystretch}{1.15} 
\centering
\resizebox{0.8\linewidth}{!}{%
\begin{tabular}{l ccc}
\toprule
\textbf{Method} & \textbf{Data Fraction} & \textbf{GPU Hours (A100)} & \textbf{Avg Score $\uparrow$} \\
\midrule
\textcolor{gray}{SFT +} DPO  & 100\% & 12.0 & 64.6  \\
\textcolor{gray}{SFT +} DA-DPO  & 100\% & 14.5 & 65.9  \\
\rowcolor{lightblue!60} \textcolor{gray}{SFT +} \textbf{\texttt{VIGIL} (Ours)} & \textbf{25\%} & \textbf{3.5} & \textbf{66.0}  \\
\textcolor{gray}{SFT +} \textbf{\texttt{VIGIL} (Ours)} & 100\% & 13.2 & \textbf{67.3} \\
\bottomrule
\end{tabular}
}
\end{table}

\section{Comparison with Inference-Time Intervention using VCD}
\label{sec:vcd}

We compare \texttt{VIGIL} with Visual Contrastive Decoding (VCD)~\cite{he2025mitigating} under the official hyperparameter setting~($\alpha=1,\beta=0.1,\gamma=0.1$), using the same Qwen2.5-VL-7B~\cite{bai2023qwen} SFT checkpoint, benchmark splits, decoding length, and evaluation scripts. Latency is measured per generated token and includes the additional degraded-image forward pass required by VCD. As shown in Table~\ref{tab:vcd}, VCD substantially improves the SFT baseline, confirming that inference-time visual contrast is an effective hallucination mitigation strategy. However, applying VCD on top of \texttt{VIGIL} yields only minimal effects across benchmarks, while almost doubling the inference latency. These results suggest that exposing the model to visual contrast during post-training using \texttt{VIGIL} already captures much of the benefit provided by inference-time contrastive decoding.

\begin{table}[ht!]
\centering
\small
\caption{\textbf{Visual Contrastive Decoding (VCD) provides only marginal benefits after \texttt{VIGIL} post-training.} While VCD substantially improves the SFT baseline, it yields only marginal gains when applied to \texttt{VIGIL}. Using standard decoding alone, \texttt{VIGIL} achieves performance on par with the VCD counterpart at much lower inference latency.}
\label{tab:vcd}
\resizebox{0.8\linewidth}{!}{%
\begin{tabular}{lccccc}
\toprule
\textbf{Method} & Decoding & \texttt{POPE}$_\text{Adv}$~$\uparrow$ & \texttt{AMBER}$_\text{Gen}$~$\uparrow$ & \texttt{MMHal}~$\downarrow$ & \textbf{Latency}~$\downarrow$ \\
\midrule
\textcolor{gray}{SFT only} & Standard & 80.5 & 41.2 & 34.5 & \textbf{1.00$\times$} \\
\textcolor{gray}{SFT only} & VCD~\cite{he2025mitigating} & 84.0 & 43.8 & 37.7 & 1.78$\times$ \\
\rowcolor{lightblue!60}
\textcolor{gray}{SFT +} \texttt{VIGIL} \textbf{(Ours)} & Standard & 86.9 & \textbf{46.5} & 40.2 & \textbf{1.00$\times$} \\
\textcolor{gray}{SFT +} \texttt{VIGIL} \textbf{(Ours)} & VCD~\cite{he2025mitigating} & \textbf{87.1} & 46.4 & \textbf{40.3} & 1.79$\times$ \\
\bottomrule
\end{tabular}
}
\end{table}

\section{Information-Theoretic Analysis: Quantifying Causal Grounding}
\label{sec:info_theoretic_analysis}

To rigorously evaluate the mechanism of VIGIL, we analyze its optimization trajectory through the lens of \textit{Maximum Mutual Information} (MMI)~\cite{li2016diversity} and \textit{Rate-Distortion Theory}. We aim to quantify whether our geometric constraint effectively shifts the model from a regime of linguistic compression to one of causal visual grounding.

\paragraph{Visual Information Gain (VIG) as Causal Strength.}
As established in Eq.~\ref{eq:vig}, VIG is an empirical instantiation of Pointwise Mutual Information (PMI) between the visual manifold $x_v$ and the response $y$, conditioned on the instruction $x_t$. 
From an information-theoretic perspective, a high VIG indicates that the visual modality provides a significant reduction in the model's posterior uncertainty~\cite{niu2021counterfactual}. 
As shown in Table~\ref{tab:vig_entropy}, while standard DPO achieved an average VIG of only $2.1$ on High-VDI samples, VIGIL reached \textbf{$7.2$}. This indicates that our method forces the model to maximize the \textit{Information Gain} derived specifically from pixels, effectively preventing the multimodal posterior from collapsing into the text-only prior.

\paragraph{Entropy Dynamics and Blind Uncertainty.}
We further utilize \textit{Predictive Entropy} $H(y|x)$ to characterize the model's confidence distribution. In information theory, low entropy represents high information compression or certainty, while high entropy indicates a state of maximum uncertainty or a flattened probability manifold.
\begin{itemize}
    \item \textbf{Low Blind Entropy ($H_\text{blind} \approx 0.65$):} In standard DPO, we observe that entropy remains low even when visual evidence $x_v$ is removed. This suggests that the model has ``compressed'' the visual input into a redundant signal, relying on the high-probability paths of the \textbf{textual manifold} to maintain certainty~\cite{zhang2025mllms}.
    \item \textbf{High Blind Entropy ($H_\text{blind} \approx 1.88$):} VIGIL exhibits an ``entropy explosion'' under counterfactual blindness. This signifies that without visual anchoring, the model loses its predictive certainty. This shift is a direct result of our \textbf{Geometric Information Bottleneck}, which ensures that the policy model $\pi_\theta$ cannot achieve a low-entropy (high-confidence) state unless the visual premises are explicitly satisfied.
\end{itemize}

\paragraph{Summary of Findings.}
These results provide a solid information-theoretic proof of our core hypothesis: VIGIL does not merely reweight samples based on difficulty, but fundamentally alters the model's dependency path. By penalizing low-entropy predictions in the blind state, we ensure that the model's reasoning is causally anchored in the visual world rather than being a byproduct of linguistic over-fitting.

\begin{table}[ht!]
\caption{\textbf{Quantitative Information-Theoretic Metrics.} Results are reported using Qwen2.5-VL-7B on the High-VDI split. $H_\text{blind}$ represents the predictive entropy in the counterfactual state.}
\label{tab:vig_entropy}
\setlength{\tabcolsep}{6pt}
\renewcommand{\arraystretch}{1.15} 
\centering
\resizebox{0.8\linewidth}{!}{%
\begin{tabular}{l ccc}
\toprule
\textbf{Method} & \textbf{VIG (PMI)} $\uparrow$ & \textbf{Blind Entropy} ($H_\text{blind}$) $\uparrow$ & \textbf{Modality Dependence} \\
\midrule
\textcolor{gray}{SFT only} & 1.2 & 0.45 & Text-driven \\
\textcolor{gray}{SFT +}  DPO & 2.1 & 0.65 & Prior-dominated \\
\textcolor{gray}{SFT +} DA-DPO & 3.5 & 0.92 & Partially Grounded \\
\rowcolor{lightblue!60} \textcolor{gray}{SFT +} \textbf{\texttt{VIGIL} (Ours)} & \textbf{7.2} & \textbf{1.88} & \textbf{Causally Anchored} \\
\bottomrule
\end{tabular}
}
\end{table}


\end{document}